\documentclass[11pt]{article}
\usepackage{iftex}
\ifPDFTeX
  \usepackage[utf8]{inputenc}
  \usepackage[T2A,T1]{fontenc}
  \usepackage{lmodern}
\else
  \usepackage{fontspec}
\fi
\usepackage[english]{babel}
\usepackage{booktabs,hyperref,amsmath,graphicx,xcolor}
\usepackage{pgfplots}
\pgfplotsset{compat=1.18}
\usepackage[margin=1in]{geometry}
\usepackage[numbers]{natbib}
\usepackage{microtype}
\newcommand{\sut}{Tuteliq Detect}
\newcommand{\sutversion}{tuteliq-detect-\allowbreak v5.20261011}

\title{SITO: Sifting a Commercial Text-Safety API for Missed Harm and False Flags in English, Ukrainian and Surzhyk\thanks{This study was commissioned by ImproveIT Solutions to provide an independent assessment of Tuteliq AI platform.}}
\author{Andrii Tuzhykov\\[2pt]
{\normalsize AI/ML Consultant / PhD Researcher}\\
{\normalsize The State Institution South Ukrainian National Pedagogical University}\\
{\normalsize named after K.\,D.~Ushynsky}\\[2pt]
{\normalsize Commissioned and funded by:}\\
{\normalsize IMPROVEIT SOLUTIONS SPÓŁKA Z OGRANICZONĄ}\\
{\normalsize ODPOWIEDZIALNOŚCIĄ (``ImproveIT Solutions'')}\\
{\normalsize\texttt{andrewtuzhykov@gmail.com}}}
\date{October 2026}

\begin{document}
\maketitle

\begin{abstract}
Platforms that host young users screen messages with commercial text-safety APIs, but independent evidence on how often such systems miss harm or flag harmless text is scarce, especially outside English, and the documentation of the one studied here leaves its false-positive rate unquantified. We evaluated one such API, Tuteliq Detect as deployed under the release identifier tuteliq-detect-v5.20261011 (not the separately published v6 model), on six harm categories: authorised push-payment fraud, money-mule recruitment, social engineering, romance scams, gambling harm and bullying. We built SITO (Synthetic Independent Text-safety benchmark with Overflagging tests; \emph{sito} is Ukrainian for `sieve'), 2,332 synthetic cases in English, Ukrainian and mixed Ukrainian, with each harmful case matched by hard harmless ones: same-topic messages, minimal-edit contrasts, warnings and retrospective accounts. On English text the system flagged 84.9--94.1\% of harmful cases and 0--4.3\% of harmless ones per category. It missed third-party reports of ongoing harm far more often than perpetrators' messages (63.6\% against 95.1\% recall, all languages), when sent without the settings the API offers for reports of harm. Ukrainian and mixed-language input was detected at least as often as English, recall was lower on text from a held-out generator family, so headline recall may be optimistic, and 3.2\% of cases received a different decision in repeated runs. A platform should measure false positives on its own traffic and check how bystander reports are handled. The datasets, generation prompts and code are released.
\end{abstract}

\section{Introduction}
Platforms that host young users increasingly rely on third-party application programming interfaces (APIs) to screen messages for fraud, exploitation and abuse. A classifier of this kind can fail in two ways. It can miss harm, or it can flag harmless messages, such as a bank's own warning, a victim describing a scam that is already over, or friends joking about a bet, and send them to moderators or block them. The second failure matters at deployment scale, because harmful messages are rare. A false-positive rate that looks small on a balanced test set can mean that most flags a moderator sees are wrong; this is the base-rate problem long known from intrusion detection \citep{axelsson2000baserate}.

Independent evidence on either failure is scarce for commercial safety APIs, and scarcer still outside English. Audits of general moderation APIs exist for hate speech \citep{hartmann2025lost}, but fraud- and child-safety-oriented detectors are mostly evaluated by their vendors on their own data. The same holds for the system we study: the vendor's published model card describes a different model version, evaluated mostly on sets authored under the same direction as its training data, states that none of its figures is a third-party evaluation and that the false-positive rate on benign text is unquantified \citep{tuteliq2026card}.

We contribute (i) SITO, six synthetic, contrastive test sets with documented generation and labelling; (ii) an evaluation of the vendor's production API through its own evaluation harness; and (iii) the first results we know of for such a system on Ukrainian and Surzhyk input.

ImproveIT Solutions, which commissioned the study, describes its motivation as follows:
\begin{quote}
``At ImproveIT Solutions, we build machine-learning systems for clients every day, and we know how much their real-world safety depends on how thoroughly they are tested. Child safety is one of the areas where the cost of failure is highest, yet there are still very few open, reproducible ways to measure how well AI systems protect children online. That is why we were glad to contribute to this research and help close that gap: a fully synthetic benchmark lets developers test their safeguards without exposing any real child's data. We hope this work helps raise the bar for how child-safety safeguards are evaluated.''

\hfill --- Ihor Dorosh, Director of Customer Success and Internal Business Operations, ImproveIT Solutions
\end{quote}

\section{Related work}
\paragraph{Contrastive testing and moderation audits.} HateCheck introduced functional tests for hate-speech classifiers, pairing hateful cases with non-hateful contrasts that defeat simple lexical rules \citep{rottger-etal-2021-hatecheck}, and was extended to ten languages \citep{rottger-etal-2022-multilingual}. XSTest applies the same idea to over-refusal, pairing 250 safe prompts with 200 unsafe contrasts \citep{rottger-etal-2024-xstest}. Contrast sets more generally probe a model's local decision boundary through small, meaningful edits \citep{gardner-etal-2020-evaluating}. We adopt minimal-edit contrasts as one of five types of harmless case. \citet{hartmann2025lost} audit five commercial moderation APIs and find both under-moderation of implicit hate and over-moderation of counter-speech and reclaimed language. Our educational and retrospective harmless cases test the analogue of counter-speech for fraud and abuse.

\paragraph{Synthetic data and small-sample statistics.} Benchmarks generated by large language models (LLMs) can follow annotation guidelines yet be easier than human-written ones and rank models differently \citep{gill2025lost}, and synthetic benchmarks can favour particular generators \citep{vanelburg2025rag}. When an LLM also judges labels, relatedness between generator and judge biases the judge \citep{li2025preference}. These findings motivate our two generator families, a control subset from a held-out third family, and a judge from a fourth. With a few hundred items, intervals based on the central limit theorem (CLT) understate uncertainty, while Wilson intervals keep nominal coverage \citep{bowyer2025clt,Wilson_1927}.

\paragraph{Fraud data and Ukrainian.} COVA-X provides synthetic multi-turn smishing conversations \citep{lochstampfor2026covax}; it contains harmful conversations only, so it cannot measure false positives. Money-mule typologies follow Europol's warning signs \citep{europol2019emma}. Gambling-harm indicators follow behavioural-marker research \citep{Delfabbro_2023}; bullying roles follow \citet{Van_Hee_2018}. Surzhyk, a Ukrainian--Russian mixed variety, has been studied sociolinguistically \citep{Hentschel_2023} and computationally for identification \citep{sira2019surzhyk}, but harmful-content detection in Surzhyk has received little attention \citep{andrusyak2018surzhyk}. Ukrainian evaluation resources for language models are appearing, for example a benchmark of ethical alignment built largely from translated scenarios \citep{kravchenko-etal-2025-ualign}; to our knowledge none targets fraud or abuse detection.

\section{System under test}
\label{sec:sut}
\sut{}, the detection system of the Tuteliq AI platform (Tuteliq AB), finds harm and fraud in text and is served through an API. The published model card \citep{tuteliq2026card} describes a model of the same line, Tuteliq Detect v6. Given a text conversation and one of 15 supported harms, among them all six of our categories, it returns a structured result: whether the harm was detected, a risk level and score, the tactics present and a recommended action, from \texttt{none} and \texttt{monitor} up to \texttt{immediate\_intervention}. English is listed as stable and Ukrainian as one of 31 beta languages; code-switched conversations are also listed as supported. The card states that the API does not run this model but a deployed detection system that is versioned separately, so we take from the card only the task and the harm taxonomy, and none of its figures.

The deployed system is identified by a string of the form \texttt{tuteliq-detect-v\allowbreak<generation>.\allowbreak<date>}, which the API reports at the start of every run. \sut{} is a fast-growing product with rapid, versioned and traceable releases: during the study new releases came out almost daily, each under its own identifier, so every result can be traced to the release that produced it; all three main runs were made on 12~October 2026 and used \texttt{\sutversion}, and what we evaluate is what the API returns under that identifier. We called the six detectors (\texttt{app-fraud}, \texttt{mule-recruitment}, \texttt{social-engineering}, \texttt{romance-scam}, \texttt{gambling-harm}, \texttt{bullying}) through the vendor's evaluation harness and its continuous-integration (CI) workflow. In the main runs we sent no language, age or setting of the conversation; multi-turn cases were sent as one text with speaker labels. An exploratory run with a declared setting is reported in Section~\ref{sec:setting}.

\section{Datasets}
\label{sec:datasets}
SITO consists of one test set per detector. Table~\ref{tab:composition} gives the composition after review: 2,332 cases, 1,179 labelled harmful and 1,153 harmless, of which 1,410 are English and 922 Ukrainian or mixed. All cases are synthetic, contain no personal data and were written for this study; none is drawn from the vendor's data.

\begin{table}[t]
\centering\small
\begin{tabular}{lrrrrrrrrr}
\toprule
Detector & Total & Harmful & en & uk & uk-par. & surzhyk & uk--ru & uk--en & translit \\
\midrule
app-fraud          & 390 & 197 & 234 & 58 & 40 & 16 & 14 & 14 & 14 \\
mule-recruitment   & 387 & 197 & 234 & 55 & 40 & 16 & 14 & 14 & 14 \\
social-engineering & 392 & 199 & 237 & 56 & 40 & 16 & 14 & 15 & 14 \\
romance-scam       & 385 & 193 & 235 & 54 & 39 & 16 & 14 & 13 & 14 \\
gambling-harm      & 391 & 194 & 236 & 58 & 39 & 16 & 14 & 14 & 14 \\
bullying           & 387 & 199 & 234 & 56 & 40 & 15 & 14 & 15 & 13 \\
\midrule
Total              & 2,332 & 1,179 & 1,410 & 337 & 238 & 95 & 84 & 85 & 83 \\
\bottomrule
\end{tabular}
\caption{Composition of the test sets after review. uk-par.: Ukrainian translations of English cases from the same set, used for paired comparison; translit: Ukrainian in Latin script.}
\label{tab:composition}
\end{table}

\paragraph{What counts as harmful.} Each set answers one question: should a moderator on a consumer or youth platform be alerted to this message? The label follows from the speaker's role (Table~\ref{tab:roles}). A message is harmful when it is itself the harmful act (the perpetrator), when a victim discloses a harm that is happening now, or when a third party reports a specific ongoing incident. It is harmless when a victim recounts an incident that is already over, when it warns or educates, or when it is ordinary communication. We separate current from past disclosure because the former calls for intervention and the latter does not. Disclosures are reported as a separate stratum, so that readers who label them differently can recompute the metrics.

\begin{table}[t]
\centering\small
\begin{tabular}{llr}
\toprule
Role or type & Label & Cases \\
\midrule
Perpetrator & harmful & 814 \\
Victim disclosure, ongoing & harmful & 258 \\
Third-party report, ongoing & harmful & 107 \\
\midrule
Same-topic legitimate message (hard negative) & harmless & 624 \\
Minimal-edit contrast of a harmful case & harmless & 171 \\
Educational or protective text & harmless & 133 \\
Victim account of a past, resolved incident & harmless & 126 \\
Unrelated everyday message (easy negative) & harmless & 99 \\
\bottomrule
\end{tabular}
\caption{Label policy and counts after review. The planned mix of harmless types was 40/25/15/10/10\%.}
\label{tab:roles}
\end{table}

\paragraph{Harmful cases.} Each category was split into subtypes taken from public typologies (money-mule recruitment: Europol's warning signs \citep{europol2019emma}; gambling harm: behavioural markers \citep{Delfabbro_2023}; bullying: participant roles \citep{Van_Hee_2018}). Each generation request fixed the subtype, the role, the channel (for example messenger, SMS, game chat or dating app), a persona and the explicitness (overt, or subtle and polite). Harmful tactics are shown at the level of indicators: no case contains working instructions, real links, real institutions' contact details or real payment identifiers.

\paragraph{Harmless cases.} Harmless cases are deliberately hard. Most share the topic, vocabulary and channel of the harmful ones: a bank's genuine warning, a real job offer, a friendly bet among adults, banter between friends. Minimal-edit contrasts keep the opening, the sender and most words of a specific harmful case but remove every harmful element (requests for money, codes or account access, pressure tied to a request, secrecy). This follows contrast sets and the functional tests of HateCheck and XSTest \citep{gardner-etal-2020-evaluating,rottger-etal-2021-hatecheck,rottger-etal-2024-xstest}.

\paragraph{Languages.} English is the primary condition. Ukrainian cases were written originally in Ukrainian rather than translated, except for the parallel subset, which translates English cases from the same set so that the two languages can be compared on the same content. Mixed-language cases cover Surzhyk, Ukrainian--Russian and Ukrainian--English code-switching, and Ukrainian written in Latin script \citep{Hentschel_2023}.

\paragraph{Generation.} Cases were generated by two model families from one written specification and one prompt template, both released: Anthropic Claude Opus~5.5 (1,191 cases) and Google Gemini~3.1~Pro (1,087). A further 54 English cases were generated by Alibaba Qwen~3.6~Plus as a held-out-family control: a family used neither for the main generation nor as the judge. Gemini and Qwen were called through a paid API gateway at temperature 1.0 in batches of five, and every prompt and raw reply was archived. The Claude share was generated in an interactive coding-assistant session from the same specification and template, and stored in the same compact format; its outputs are released, but its prompts were not archived call by call, so this share cannot be regenerated as the others can. We had planned a human-written control subset; it was replaced by the held-out-family control for lack of time, so all data in this study are synthetic (Section~\ref{sec:limitations}).

\paragraph{Quality control and review.} Automatic checks covered format, script and language per group, personal data (only reserved example domains and invalid number ranges are allowed), and shortcut cues such as differences in length between harmful and harmless cases. An LLM judge from a fourth family, OpenAI GPT-5.4~mini, labelled every case without seeing the label, role or type. A first pass showed that many Gemini minimal-edit contrasts still contained the harmful element (38 of 71 judged harmful, against 4 of 90 for Claude). We sharpened the contrast instruction and regenerated those cases. The author then reviewed every case on which the judge disagreed with the label or was unsure, a random 20\% of the Ukrainian and mixed-language cases, and all held-out-family cases, and recorded a decision for every case in the six sets. The review excluded 68 cases: 43 as unrealistic, 22 as ambiguous, 2 as mislabelled and 1 as written in the wrong language. On the final sets the judge agrees with the labels on 96.7\% of cases (Cohen's $\kappa = 0.935$; 4 cases judged unsure). The 76 remaining disagreements were kept after review and are released with the judge's verdicts. No second human annotator was available, which we list as a limitation.

\paragraph{Release.} We release the sets, the specification, the prompts, the raw generations, the review decisions with reasons and the code under CC BY 4.0 (see Data and code availability). Each set is a JSON Lines file in the evaluator format of the vendor's harness, with the fields \texttt{id}, \texttt{label}, \texttt{text} or \texttt{messages}, \texttt{detector} and \texttt{group}. Generator, role, type, channel and persona are included for stratified analysis but are not sent to the API. The harness passes a conversation to these detectors as one text, each turn prefixed by its speaker label (\texttt{adult} or \texttt{unknown}). A \texttt{child} label would tell the system that a minor is present, which no other field does, so the 77 conversations written with a child speaker are sent with that speaker as \texttt{unknown} and are marked in the released data.

\section{Method}
\label{sec:method}
\paragraph{Decision rule.} A case counts as flagged when the API recommends an action on it, such as review by a moderator or blocking. When the API sees no risk, or only records the message without asking anyone to act, the case is not flagged.

\paragraph{Runs.} Each run sends every case of the six sets once through the vendor's CI workflow. Three runs were made, one after another, on the same system identifier. Run~1 is the primary run; runs 2 and 3 measure stability and give the range over runs in Appendix~\ref{app:stability}. Appendix~\ref{app:repro} describes how to reproduce the runs.

\paragraph{Metrics.} For each detector we report recall (the share of harmful cases flagged) and the false-positive rate (FPR; the share of harmless cases flagged), with 95\% Wilson confidence intervals (CIs), which keep their nominal coverage at these sample sizes where normal-approximation intervals do not \citep{Wilson_1927,bowyer2025clt}; precision, accuracy and F1 on the balanced sets; the area under the receiver operating characteristic curve (ROC-AUC) of the API's risk score, with a 1,000-resample bootstrap interval; and precision at assumed prevalences of 1\%, 5\% and 10\%, computed from recall and false-positive rate. The last matters most for deployment: on a balanced test set precision says little about how many flags a moderator will see that are wrong when harm is rare.

\paragraph{Questions.} The primary question (Q1) is recall and false-positive rate per category on English text. Secondary questions: Q2, differences between English and the Ukrainian and mixed-language groups, with an exact McNemar test on correctness for the English--Ukrainian translation pairs; Q3, which kinds of harmless text are flagged, including the share of minimal-edit contrast pairs on which both cases are classified correctly; Q4, differences between generators, which would indicate an artefact of synthetic data; and Q5, stability, as the share of cases whose decision differs between runs and Fleiss' $\kappa$ across runs. Analyses by subtype, explicitness and format are exploratory.

\section{Results}
\label{sec:results}
\paragraph{Primary result (Q1).} Table~\ref{tab:primary} gives recall and false-positive rate per category on English text. Recall on English harmful cases ranged from 84.9\% (app-fraud) to 94.1\% (mule recruitment), 88.6\% pooled; the false-positive rate ranged from 0\% (app-fraud, romance scam, gambling harm) to 4.3\% (social engineering), 1.3\% pooled. The system therefore errs towards missing harm rather than flagging harmless text: about one English harmful case in nine was not flagged. Where no false positive was observed, precision at low prevalence is bounded only by the upper end of the false-positive interval; Table~\ref{tab:primary} gives that conservative bound. Table~\ref{tab:balanced} gives precision, accuracy and F1 on the balanced sets and ROC-AUC with its interval. In gambling harm the risk score separated the classes perfectly (ROC-AUC 1.000) while the action-based rule missed 14.4\% of harmful cases.

\begin{table}[t]
\centering\small\setlength{\tabcolsep}{4pt}
\begin{tabular}{lrcccc}
\toprule
Detector & $n$ & Recall [95\% CI] & FPR [95\% CI] & Precision bound @1 / 5 / 10\% & ROC-AUC \\
\midrule
app-fraud          & 234 & 84.9 [77.4, 90.2] & 0.0 [0.0, 3.2] & 19.5 / 55.7 / 72.7 & 0.959 \\
mule-recruitment   & 234 & 94.1 [88.3, 97.1] & 1.7 [0.5, 6.1] & 12.8 / 43.4 / 61.8 & 0.997 \\
social-engineering & 237 & 92.5 [86.4, 96.0] & 4.3 [1.8, 9.6] & 8.3 / 32.1 / 49.9 & 0.952 \\
romance-scam       & 235 & 85.5 [78.0, 90.7] & 0.0 [0.0, 3.2] & 20.0 / 56.5 / 73.3 & 0.985 \\
gambling-harm      & 236 & 85.6 [78.1, 90.8] & 0.0 [0.0, 3.2] & 20.0 / 56.6 / 73.4 & 1.000 \\
bullying           & 234 & 89.1 [82.2, 93.5] & 1.7 [0.5, 6.1] & 11.9 / 41.4 / 59.9 & 0.960 \\
\bottomrule
\end{tabular}
\caption{Primary result: English cases, run~1. Intervals are 95\% Wilson intervals; the range over the three runs is given in Appendix~\ref{app:stability}. Precision at prevalence is the conservative bound computed from the lower recall limit and the upper FPR limit; the point estimates are 100\% wherever no false positive was observed. All values in per cent except ROC-AUC.}
\label{tab:primary}
\end{table}

\begin{table}[t]
\centering\small
\begin{tabular}{lcccc}
\toprule
Detector & Precision & Accuracy & F1 & ROC-AUC [95\% CI] \\
\midrule
app-fraud          & 100.0 & 92.3 & 0.918 & 0.959 [0.932, 0.981] \\
mule-recruitment   & 98.2 & 96.2 & 0.961 & 0.997 [0.992, 1.000] \\
social-engineering & 95.7 & 94.1 & 0.941 & 0.952 [0.918, 0.981] \\
romance-scam       & 100.0 & 92.8 & 0.922 & 0.985 [0.966, 0.998] \\
gambling-harm      & 100.0 & 92.8 & 0.922 & 1.000 [1.000, 1.000] \\
bullying           & 98.1 & 93.6 & 0.934 & 0.960 [0.933, 0.982] \\
\bottomrule
\end{tabular}
\caption{Metrics on the balanced sets: English cases, run~1. Precision and accuracy in per cent. On a balanced set precision overstates deployment precision (Table~\ref{tab:primary}). The ROC-AUC interval is a 1,000-resample bootstrap.}
\label{tab:balanced}
\end{table}

\paragraph{Languages (Q2).} Ukrainian and mixed input did not lower recall: every Ukrainian and mixed group had recall at or above the English 88.6\% (Table~\ref{tab:languages}). On the 238 English--Ukrainian translation pairs, 4 were right in English and wrong in Ukrainian and 7 the reverse (exact McNemar $p=0.55$). The false-positive rate on original Ukrainian text was higher than on English (4.3\% against 1.3\%), with overlapping intervals. All 7 false positives on original Ukrainian text were minimal-edit contrasts (7 of 36, against 3 of 111 in English), and none of the 12 translated contrasts was flagged.

\begin{table}[t]
\centering\small\setlength{\tabcolsep}{4pt}
\begin{tabular}{lrccc}
\toprule
Group & $n$ & Recall [95\% CI] & FPR [95\% CI] & ROC-AUC [95\% CI] \\
\midrule
English          & 1,410 & 88.6 [86.1, 90.7] & 1.3 [0.7, 2.4] & 0.975 [0.966, 0.983] \\
Ukrainian        & 337   & 92.6 [87.8, 95.6] & 4.3 [2.1, 8.7] & 0.975 [0.956, 0.990] \\
Ukrainian (translation of English) & 238 & 94.1 [88.3, 97.1] & 0.8 [0.1, 4.6] & 0.988 [0.970, 1.000] \\
Surzhyk          & 95    & 97.9 [89.1, 99.6] & 0.0 [0.0, 7.6] & 0.998 [0.993, 1.000] \\
Ukrainian--Russian & 84  & 95.2 [84.2, 98.7] & 0.0 [0.0, 8.4] & 0.986 [0.956, 1.000] \\
Ukrainian--English & 85  & 100.0 [91.6, 100.0] & 2.3 [0.4, 12.1] & 0.998 [0.990, 1.000] \\
Latin script     & 83    & 97.6 [87.7, 99.6] & 0.0 [0.0, 8.6] & 0.998 [0.992, 1.000] \\
\bottomrule
\end{tabular}
\caption{Results by language group, pooled over the six categories, run~1, in per cent with 95\% Wilson intervals. Small groups have wide intervals.}
\label{tab:languages}
\end{table}

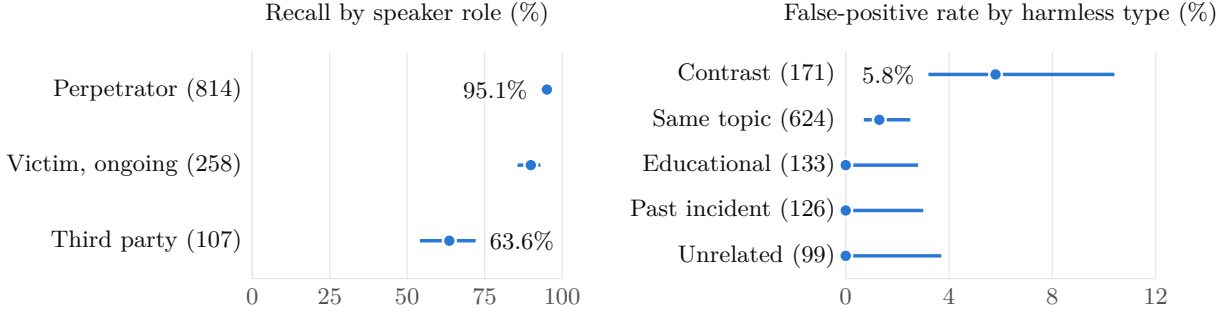
\begin{figure}[t]
\centering
\definecolor{sitoseries}{HTML}{2A78D6}\definecolor{sitoink}{HTML}{0B0B0B}\definecolor{sitomuted}{HTML}{52514E}\definecolor{sitogrid}{HTML}{E4E3DF}
\begin{tikzpicture}
\begin{axis}[name=roles, scale only axis, width=4.1cm, height=3.0cm, xmin=0, xmax=100, ymin=0.5, ymax=3.5, xtick={0,25,50,75,100}, ytick={1,2,3}, yticklabels={{Third party (107)},{Victim, ongoing (258)},{Perpetrator (814)}}, yticklabel style={font=\footnotesize, text=sitoink}, xticklabel style={font=\footnotesize, text=sitomuted}, xmajorgrids, grid style={draw=sitogrid, line width=0.4pt}, axis line style={draw=sitogrid}, axis x line*=bottom, axis y line*=left, tick style={draw=none}, title={Recall by speaker role (\%)}, title style={font=\footnotesize, text=sitoink, align=center}]
\addplot+[only marks, mark=*, mark size=2.4pt, mark options={fill=sitoseries, draw=white, line width=0.9pt},
  error bars/.cd, x dir=both, x explicit, error bar style={draw=sitoseries, line width=1.3pt}, error mark=none]
  table[x=v, y=pos, x error plus expr=\thisrow{hi}-\thisrow{v}, x error minus expr=\thisrow{v}-\thisrow{lo}] {
    pos v lo hi
    1 63.6 54.1 72.1
    2 89.9 85.6 93.0
    3 95.1 93.4 96.4
  };
  \node[anchor=west, font=\footnotesize, text=sitoink] at (axis cs:72.1,1) {\,63.6\%};
  \node[anchor=east, font=\footnotesize, text=sitoink] at (axis cs:93.4,3) {95.1\%\,};
\end{axis}
\begin{axis}[at={(roles.outer east)}, anchor=outer west, xshift=0.4cm, scale only axis, width=4.1cm, height=3.0cm, xmin=0, xmax=12, ymin=0.5, ymax=5.5, xtick={0,4,8,12}, ytick={1,2,3,4,5}, yticklabels={{Unrelated (99)},{Past incident (126)},{Educational (133)},{Same topic (624)},{Contrast (171)}}, yticklabel style={font=\footnotesize, text=sitoink}, xticklabel style={font=\footnotesize, text=sitomuted}, xmajorgrids, grid style={draw=sitogrid, line width=0.4pt}, axis line style={draw=sitogrid}, axis x line*=bottom, axis y line*=left, tick style={draw=none}, title={False-positive rate by harmless type (\%)}, title style={font=\footnotesize, text=sitoink, align=center}]
\addplot+[only marks, mark=*, mark size=2.4pt, mark options={fill=sitoseries, draw=white, line width=0.9pt},
  error bars/.cd, x dir=both, x explicit, error bar style={draw=sitoseries, line width=1.3pt}, error mark=none]
  table[x=v, y=pos, x error plus expr=\thisrow{hi}-\thisrow{v}, x error minus expr=\thisrow{v}-\thisrow{lo}] {
    pos v lo hi
    1 0.0 0.0 3.7
    2 0.0 0.0 3.0
    3 0.0 0.0 2.8
    4 1.3 0.7 2.5
    5 5.8 3.2 10.4
  };
  \node[anchor=east, font=\footnotesize, text=sitoink] at (axis cs:3.2,5) {5.8\%\,};
\end{axis}
\end{tikzpicture}
\caption{Recall by speaker role (left) and false-positive rate by type of harmless case (right), all languages, run~1, no setting declared, with 95\% Wilson intervals; numbers of cases in parentheses. Third party: third-party report of ongoing harm; victim, ongoing: victim disclosure of ongoing harm; same topic: same-topic legitimate message; contrast: minimal-edit contrast; educational: educational or protective text; past incident: victim account of a past incident; unrelated: unrelated everyday message (Table~\ref{tab:roles}).}
\label{fig:roles}
\end{figure}

\paragraph{Which harmless text is flagged (Q3).} False positives concentrated on minimal-edit contrasts (10 of 171, 5.8\%) and, less often, on same-topic legitimate messages (8 of 624, 1.3\%); none of the 133 educational texts, 126 retrospective accounts or 99 unrelated messages was flagged. Both members of a contrast pair were classified correctly in 139 of 149 pairs (93.3\%). Across all languages, recall depended on who was speaking (Figure~\ref{fig:roles}): 95.1\% [93.4, 96.4] on perpetrator messages (774 of 814), 89.9\% [85.6, 93.0] on ongoing victim disclosures (232 of 258) and 63.6\% [54.1, 72.1] on third-party reports (68 of 107), an interval well below the other two. Like all main-run cases, the reports were sent with no declared setting (Section~\ref{sec:setting}). Recall was also lower on subtle than on overt cases (88.2\% against 93.6\%) and on conversations than on single messages (87.7\% against 92.7\%); conversations also had the higher false-positive rate (3.1\% against 1.0\%). The lowest-recall subtypes (exploratory, 21 to 37 cases each) were love bombing (58.8\%), rumours and relational bullying (66.7\%), moving the conversation off the platform (70.4\%) and loss of control over gambling (75.0\%).

\paragraph{Generators (Q4).} Recall was 93.6\% [91.4, 95.3] on cases generated by Claude, 89.5\% [86.7, 91.8] on Gemini and 67.9\% [49.3, 82.1] on the 54-case Qwen control (19 of 28 harmful cases); the false-positive rates were 0.7\%, 2.3\% and 7.7\% (2 of 26). The Qwen interval does not overlap Claude's. The control is small and English-only, but the direction is the one a synthetic-data artefact would produce: text from the generator families that produced most of the set was easier for the system than text from a family held out of it.

\paragraph{Stability (Q5).} The three runs used the same system identifier and returned no failed call. Of the 2,332 cases, 74 (3.2\%) received a different decision in at least one run; Fleiss' $\kappa$ over the three runs was 0.957. On English text, recall per category moved by at most 3.4 points between runs (bullying, 85.7--89.1\%; romance scam, 82.1--85.5\%) and the false-positive rate by at most 0.8 points (romance scam, one harmless case in run~2; Appendix~\ref{app:stability}).

\section{Declared setting (exploratory)}
\label{sec:setting}
The API accepts an optional declared setting of a conversation (\texttt{conversation\_type}). The vendor's documentation, updated on 12~October, gives guidance on choosing it: one value per surface, a role-bearing setting such as \texttt{helpline} only where that role is really present, and none for SMS, e-mail or phone transcripts. The same guidance states that a declared setting can lower the urgency of an answer and never makes the API flag a message it would not flag otherwise. We made one further full run on copies of the six sets in which every case carries a setting taken only from its channel by a fixed table (Table~\ref{tab:setting}, note). IDs, labels and text are unchanged. The run (13~October 2026) scored all 2,332 cases with no failed call. It reported \texttt{tuteliq-detect-\allowbreak v5.20261012}, a release with the same model as \texttt{v5.20261011} and a changed API layer, the layer that turns a request into the model's input and the model's output into the response, which now passes the declared setting on to the model. We compare it with run~3 on the same cases.

\begin{table}[t]
\centering\small\setlength{\tabcolsep}{4pt}
\begin{tabular}{lrcccc}
\toprule
Cases & $n$ & \multicolumn{1}{c}{Harmful} & \multicolumn{1}{c}{Harmless} & Recall & FPR \\
 & & lost / gained & lost / gained & & \\
\midrule
\multicolumn{6}{l}{\emph{Run 3 $\rightarrow$ run with setting (\texttt{v5.20261011} $\rightarrow$ \texttt{v5.20261012})}} \\
All                         & 2,332 & 16 / 61 & 3 / 13 & 90.5 $\rightarrow$ 94.3 & 1.6 $\rightarrow$ 2.4 \\
No setting declared         & 443   & 2 / 18  & 1 / 3  & 90.0 $\rightarrow$ 97.3 & 2.2 $\rightarrow$ 3.1 \\
Setting declared            & 1,889 & 14 / 43 & 2 / 10 & 90.6 $\rightarrow$ 93.6 & 1.4 $\rightarrow$ 2.3 \\
\quad of which \texttt{helpline} & 144 & 7 / 1 & 0 / 0  & 92.9 $\rightarrow$ 87.6 & 3.2 $\rightarrow$ 3.2 \\
\midrule
\multicolumn{6}{l}{\emph{Reference: run 2 $\rightarrow$ run 3 (same identifier, no setting)}} \\
All                         & 2,332 & 18 / 22 & 3 / 2  & 90.2 $\rightarrow$ 90.5 & 1.6 $\rightarrow$ 1.6 \\
Channels left undeclared    & 443   & 4 / 2   & 0 / 1  & 90.9 $\rightarrow$ 90.0 & 1.8 $\rightarrow$ 2.2 \\
\bottomrule
\end{tabular}
\caption{The run with a declared setting compared with run~3 on the same cases, all languages. ``Lost'' counts cases flagged in the earlier run and not in the later one; ``gained'' the reverse. Recall and FPR in per cent. Setting by channel: \texttt{direct\_message} (one-to-one messengers, marketplace chat; 919 cases), \texttt{group\_chat} (group, family, parents' and class chats, Discord; 453), \texttt{public\_post} (posts, comments, forums, job boards; 282), \texttt{helpline} (144), \texttt{dating} (42), \texttt{gaming\_chat} (20), \texttt{workplace} (17), \texttt{education} (school newsletter; 12); none for SMS, e-mail, phone transcripts and pop-up text (443).}
\label{tab:setting}
\end{table}

\paragraph{Result.} Declaring the setting did not raise detection; it lowered alerts only on helpline conversations. Recall over all cases rose from 90.5\% to 94.3\% and the false-positive rate from 1.6\% to 2.4\% (Table~\ref{tab:setting}). The rise also appeared, and was larger, on the 443 cases that declared no setting and sent the same request as in run~3: 18 harmful cases gained a flag and 2 lost one, whereas between runs 2 and 3 on one identifier the same cases moved 2 and 4. We therefore attribute the overall rise to the change of the API layer between the two releases, not to the setting. On \texttt{helpline} conversations the direction reversed: 7 harmful cases lost their flag and 1 gained one (recall 92.9\% to 87.6\%); 4 of the 7 were third-party reports. This matches the vendor's guidance. On the other settings, flags were gained, as on the undeclared control, or did not change. Because the release changed together with the setting, the effect of the setting alone cannot be separated from that of the release on cases that declared one; the run lies outside Q1--Q5 and is reported descriptively.

\section{Discussion}
\paragraph{False positives at deployment scale.} On deliberately hard harmless text the system rarely flagged anything: 1.3\% of English harmless cases, and none of the educational texts, retrospective victim accounts or unrelated messages. What that means for a moderation queue depends on prevalence. At 1\% prevalence, the observed rates put the share of correct flags at 18--36\% in the three categories with any false positive (social engineering, bullying, mule recruitment). In the other three no harmless case was flagged, but with 115 to 118 harmless English cases per category the data cannot exclude a false-positive rate of 3.2\%, and at that rate only one flag in five would be correct (Table~\ref{tab:primary}). These figures come from harmless text built to resemble harm, so they are a pessimistic reading for ordinary traffic and an optimistic one for platforms whose traffic is mostly about money, dating or games. The model card gives no false-positive figure for any of these six categories \citep{tuteliq2026card}; ours are an external measurement, and still too few to fix a deployment precision: that needs a platform's own traffic.

\paragraph{Where harm is missed.} About one English harmful case in nine was not flagged, and the misses were not spread evenly. Third-party reports of ongoing harm were missed in more than a third of cases (63.6\% recall), against 4.9\% of perpetrator messages; a platform that relies on bystanders reporting what they see would lose many of those reports. These figures describe reports sent with no declared setting. The API also offers settings for reports of harm, for example to a fraud desk, an abuse-reporting channel or a moderator, and handles such reports differently; we did not test them. Subtle cases were missed more often than overt ones, and the weakest subtypes were those in which the harmful act is not yet a request: love bombing, rumours and relational bullying, moving a conversation off the platform, and loss of control over gambling. In gambling harm the risk score separated harmful from harmless cases perfectly while the recommended action missed 14.4\% of harmful cases, so in that category the misses lie in how the score is turned into an action rather than in the score itself. In the exploratory run a declared setting did not raise detection; on helpline conversations it removed flags, including from third-party reports (Section~\ref{sec:setting}).

\paragraph{Ukrainian and mixed language.} Ukrainian, Surzhyk, code-switched and transliterated input was detected at least as often as English, and on translated pairs the two languages did not differ. The vendor lists Ukrainian as a beta language and does not mention Surzhyk \citep{tuteliq2026card}, so this is a better result than the documentation promises. The false-positive rate on original Ukrainian text was, however, higher than on English (4.3\% against 1.3\%; the intervals overlap). All of those false positives were minimal-edit contrasts written in Ukrainian, while the Ukrainian translations of English cases, contrasts included, had a rate of 0.8\%; the difference therefore more likely lies in these contrasts than in the language itself. With 36 such contrasts we cannot tell whether they kept more of the harmful cue than intended or the system reads Ukrainian contrasts less well, and a platform whose Ukrainian traffic resembles them could still see more wrong flags. The mixed-language groups are small (83--95 cases) and their intervals wide.

\paragraph{Synthetic data and generator family.} Recall was 93.6\% on cases written by Claude, 89.5\% on Gemini and 67.9\% on the 54-case Qwen control. The control is small and English-only, and the difference may reflect how each family writes rather than any familiarity of the system with a style. Either way, our headline recall may be optimistic, and a synthetic benchmark built from one or two generator families should report results by generator and include a held-out family, as we did, or human-written text, as we planned and could not do.

\paragraph{What an integrating platform should test.} Four things follow for a platform that integrates such an API. First, measure the false-positive rate on a sample of its own traffic, since no test set fixes precision at its prevalence. Second, check the roles its users write in, in particular bystander reports and disclosures, and declare a setting deliberately: \texttt{helpline} lowers alerts, also on reports passed on by a third party, which suits a helpline with a counsellor in the conversation but not a platform that relies on the API to escalate such reports. Third, do not assume a single call is final: 3.2\% of our cases changed decision between identical runs on the same system identifier, so borderline messages can pass on one call and be flagged on the next. Fourth, record the system identifier with every decision, because new releases come out often (Section~\ref{sec:sut}). Our figures hold for the deployed release \texttt{\sutversion}, not for the published v6 model or for later releases.

\section{Limitations}
\label{sec:limitations}
\paragraph{Synthetic data only.} All cases were written by language models. Synthetic test sets can be easier than human-written ones and can favour particular generators \citep{gill2025lost,vanelburg2025rag}. We used two generator families, a held-out third family and a judge from a fourth, but a planned human-written control subset was not produced, so we cannot measure the gap to human text directly. Recall on the held-out Qwen control was 67.9\%, against 93.6\% and 89.5\% on the two main families (Section~\ref{sec:results}), so headline recall may be optimistic for human-written text; a human-written test set is the next validation step.

\paragraph{One human annotator.} Labels follow from a written role policy and were checked by an LLM judge on every case and by the author on judge disagreements, a 20\% sample of the Ukrainian and mixed-language cases and all held-out-family cases. No second human annotator was available, so we cannot report human inter-annotator agreement. A next version should add a second annotator, ideally a native speaker of Ukrainian familiar with Surzhyk.

\paragraph{One system, no baseline.} We evaluate six detectors of one system, on text only, and compare them with no other system. The figures describe this system under the identifier \texttt{\sutversion} and say nothing about other vendors; the released sets can be run against any other API or model. The LLM judge is not a baseline, because the labels were reviewed against its verdicts. Bystander reports were tested without the settings the API offers for reports of harm.

\paragraph{Dependent cases.} The intervals treat cases as independent. They are not fully so: each minimal-edit contrast is built from a harmful case, and the Gemini and Qwen cases were generated in batches of five from shared requests. The intervals may therefore be somewhat too narrow; resampling by pair or batch would give a more conservative estimate.

\section*{Competing interests}
This study was commissioned and funded by ImproveIT Solutions, which was engaged for it by Tuteliq AB, the vendor of the evaluated system. Tuteliq provided, free of charge, access to its production API, an API key and its evaluation harness. Tuteliq saw the manuscript only to correct factual errors and had no editorial control. The evaluation was conducted externally, and its design, analysis and conclusions are the author's; this is the sense in which we call it independent. The author received no payment from Tuteliq.

\section*{Use of generative AI}
Following arXiv policy, we report that text-to-text generative AI was used to (a) generate the synthetic test cases (Anthropic Claude Opus~5.5, Google Gemini~3.1~Pro and Alibaba Qwen~3.6~Plus; model identifiers and prompts are released), (b) relabel every case blind as an independent check (OpenAI GPT-5.4~mini), and (c) assist with code and drafting (Anthropic Claude). No response of the evaluated API was given to any generative model. The author reviewed all outputs and takes full responsibility for the content.

\section*{Data and code availability}
The SITO sets, the specification, the prompts and raw generations, the review decisions and the code for generation, quality control, review and analysis are available on Hugging Face at \url{https://huggingface.co/datasets/Improve-IT-Solutions/SITO} (DOI \href{https://doi.org/10.57967/hf/10883}{10.57967/hf/10883}, revision \texttt{d9fdee3}), with access granted on request, under the Creative Commons Attribution 4.0 licence (CC BY 4.0); ImproveIT Solutions holds the rights. The licence permits any use; we ask that the sets be used for evaluation and not for training, since a test set that has been trained on no longer measures anything. Each file carries a canary string so that inclusion in training data can be detected. Per-case predictions are not released, because the API's responses are the vendor's confidential information; the aggregate figures are given in this paper. This paper, the revision above and the release identifiers \texttt{\sutversion} (main runs) and \texttt{tuteliq-detect-\allowbreak v5.20261012} (Section~\ref{sec:setting}) together identify the evidence reported here.

\bibliographystyle{plainnat}
\bibliography{refs}

\appendix
\section{Stability across runs}
\label{app:stability}
\begin{table}[h]
\centering\small
\begin{tabular}{lcc}
\toprule
Detector & Recall, runs 1 / 2 / 3 & FPR, runs 1 / 2 / 3 \\
\midrule
app-fraud          & 84.9 / 84.0 / 84.9 & 0.0 / 0.0 / 0.0 \\
mule-recruitment   & 94.1 / 94.1 / 95.8 & 1.7 / 1.7 / 1.7 \\
social-engineering & 92.5 / 91.7 / 92.5 & 4.3 / 4.3 / 4.3 \\
romance-scam       & 85.5 / 82.9 / 82.1 & 0.0 / 0.8 / 0.0 \\
gambling-harm      & 85.6 / 87.3 / 87.3 & 0.0 / 0.0 / 0.0 \\
bullying           & 89.1 / 85.7 / 86.6 & 1.7 / 1.7 / 1.7 \\
\bottomrule
\end{tabular}
\caption{English cases, per run (per cent). Over all 2,332 cases, 74 (3.2\%) changed decision in at least one run; Fleiss' $\kappa=0.957$. No call failed in any run.}
\label{tab:stability}
\end{table}

\section{Reproducing the runs}
\label{app:repro}
The released sets and code are enough to repeat the evaluation with access to the API. First, send each case once to the detector of its set: a single message as its text, a conversation as one text in which each turn is prefixed by its speaker label, with no language, age or setting declared. Second, record whether the API recommends an action and the risk score it returns; a case counts as flagged when an action is recommended (Section~\ref{sec:method}), and a call that fails is excluded and counted. Third, compute the metrics with the released analysis code, which takes one record per case with its id, label, flag and risk score. A later run measures a later system: new releases come out often (Section~\ref{sec:sut}), and 3.2\% of cases changed decision even between identical runs (Section~\ref{sec:results}), so a repeat should be compared on one identifier and read against that spread. The copies with a declared setting (Section~\ref{sec:setting}) can be rebuilt from each case's channel with the mapping in the note to Table~\ref{tab:setting}.
\end{document}